\documentclass[letterpaper]{article}
\usepackage{aaai}
\usepackage{times}
\usepackage{helvet}
\usepackage{courier}
\usepackage{graphicx}
\begin{document}
%
\title{Decision-support for the Masses by Enabling Conversations with Open Data}
\author{Biplav Srivastava\\
IBM \\
}
\maketitle
\begin{abstract}
\begin{quote}
Open data refers to data that is freely available for reuse. Although there has been rapid increase in availability of open
data to public in the last decade, this has not translated into better decision-support tools for them.
We propose intelligent conversation generators  as a grand challenge that would automatically create data-driven conversation interfaces (CIs), also known as chatbots or dialog systems, from open data and deliver personalized analytical insights to users based on their contextual needs.  Such generators will not only help bring Artificial Intelligence (AI)-based solutions for important societal problems to the masses but also advance AI by providing an integrative testbed for human-centric AI and filling gaps in the state-of-art towards this aim.


\end{quote}
\end{abstract}
      

\section{Introduction}
\label{sec:intro}

As the world has shifted towards an increased digital economy and 
organizations have adopted the open data principles to make their data widely available for reuse \cite{od-tutorial,open-data-state-report}, there is an unprecedented opportunity to generate insights to improve the conditions of people around the world towards basic concerns of living like health, water, energy, traffic, community and environment \cite{ai-opendata-tutorial}. However,  
the current  situation is that available interfaces of search \footnote{https://toolbox.google.com/datasetsearch} and visualization  for open data \footnote{Examples - http://data.gov, http://data.gov.in.} are targeted towards developers and are rudimentary when compared to what people want - relevant, prescriptive, data-driven insights for taking decisions. 
Building intelligent interfaces with open data needs specialization in AI and data management, and is costly and consequently, slow to develop. 
The first glimpses can be seen in chatbot released by the state of Kansas \cite{kansas-chatbot} in the US while the private sector has tried to provide a natural question-answering interface \cite{usa-facts} to US data. The research community has also started to build many prototypes in the area \cite{wa-aaai18,tyson-aaai18}, and we seek to accelerate the phenomenon.


\begin{table}[t!]
\scriptsize
\centering
\begin{tabular}{|l|l|l|l|l|}
\hline
\textbf{S.No.} & \textbf{Dimension} & \textbf{Variety} \\ \hline
1 & User & 1, multiple \\ \hline
2 & Modality &  only conversation,  only speech, \\ 
  &   & multi-modal (with point, map, ...) \\ \hline
3 & Data source & none, static, dynamic \\ \hline
4 & Personalized &  no, yes \\ \hline
5 & Form & virtual agent, physical device, robot \\ \hline
6 & Purpose & socialize, goal: information seeker,  \\ 
  &  & goal: action delegate \\ \hline
7 &  Domains & general, health, water, traffic, ...\\ \hline

\end{tabular}
\caption{Different Types of Conversation Interfaces}
\label{tab:chatbot-types}
\end{table}

For people, natural interfaces to data have long been known to be effective.  Previous attempts to build natural interfaces to data include building question answer systems directly from 
data\cite{neural-qa,sem-parsing,ie-freebase,qa-knowledgebase} and providing natural interfaces to a query language\cite{nlidb-intro}. But they support a one-off interaction with data while conversation is inherently iterative.

A conversation interface (CI), also known as chatbot or dialog system\footnote{Some researchers use the term chatbot exclusively for agents that perform chit-chat. Instead, we use the terms conversation agents, chatbots and dialog systems interchangeably to mean task-oriented conversation agents which is the focus of this paper.} \cite{dialog-intro}, is an automated agent, whether physical like robots or abstract like an avatar, that can not only interact with a person, but also  take actions on their behalf and get things done. A simple taxonomy of interfaces we consider is shown in Table~\ref{tab:chatbot-types}. One can talk to a chatbot or, if speech is not supported, type an input and get the system’s response. They can be embedded along with other interaction modalities to give a rich user experience. The chatbot may converse without a goal in pleasantries and hence not need access to data sources, or be connected to a static data source like a company directory or a dynamic data source like weather forecast. The application scenarios become more  compelling when the chatbot works in a dynamic environment, e.g., with sensor data,  interacts with groups of people who come and go rather than only an individual at a time, and adapts its behavior to peculiarities of user(s). 

There has been a surge in availability of chatbots for people on mobile phones and physical devices like the Amazon Alexa and Google Home. Numerous platforms have emerged to create them quickly for any domain \cite{chatbot-survey-accenture}.
Chatbots have been deployed in customer care in many industries where they are expected to save over \$8 billion per annum by 2022 \cite{chatbot-cc-juniper}. Chatbots can help users especially in unfamiliar domains when users do  not know everything about the data, its metadata, kinds of analyses possible and the implications of using the insights. 

However, the process to build chatbots needs long development cycle and is costly. The main problem here is dialog management, i.e., creating dialog responses to user's utterances by trying to understand the 
user's intent, determining the most suitable response, building queries to retrieve data and deciding the
next course of action (responding, seeking more information
or deferring). The chatbot so created also has to be tested for functional and non-function characteristics, and social behavior.


We envision a simpler and cost-efficient process where the user can point to a data-source like water quality and regulations, and gets a chatbot custom-generated so that the user can ask whether they can safely drink a location's water. The system could also explain its recommendation and justify if new information is provided about water usage on other days or locations. 
The user can then point to a disease data-source and the chatbot automatically updates itself so that it can now converse and answer questions about diseases, in general, but also water-based diseases for a specific location, in particular. Changing domains, the user can point to travel data about events happening in their favorite cities and get a new chatbot to advise on their upcoming vacation. 
Conversation agents in these and other domains are 
increasingly available but as noted earlier, they are 
costly and time-consuming to build from scratch.

We envisage software programs, that we call {\em chatbot generators}, for generating conversation interfaces to deliver data-driven insights and also become personalized over time and data sources. The chatbot generator would be able to quickly adapt to new domains and data sources based on a user's needs, generate chatbots that are broadly useful, trustable by being able to transparently explain their decision process and aware of fairness issues, and able to deploy chatbots  widely in different forms (Table~\ref{tab:chatbot-types}).

We now discuss open data, challenges in using them and some use-cases where insights from them can help people. Next, we review how conversation can help 
address the challenges and how the
proposed chatbot generator can fill the gap for common
usage patterns. We will
use water as a case-study throughout to motivate how general people may benefit, where a prototypical multi-modal chatbot called  {\em Water Advisor} (WA) was recently described 
\cite{wa-aaai18}. 
We identify AI opportunities for 
learning, reasoning, representation and execution, along with human-centric design and ethics, to motivate more  conversation applications.

\section{Open Data and Its Challenges}
\label{sec:open-data}

Open data is an important trend in governance and computing over the last decade \cite{open-data-state-report,odb-4ed}. According to {\em Open Data Catalogs}, 
there are over 550  repositories, each with data items (called resources) ranging from tens to thousands, spanning most areas of human endeavor\footnote{https://datacatalogs.org/, Accessed 28-Dec-2018.}. The Open Data Barometer \cite{odb-4ed} prepared a report surveying 1,725 datasets from 15 different sectors across 115 countries. In the first wave of the trend, which started around 2008, the focus was on acquiring data and making it available on scalable public platforms. In the second wave, the focus has been on driving consumption by providing richer semantics \cite{open-data-india-report}. 

Many use-cases have been published demonstrating how insights 
from open data using AI methods can benefit people.
For example, \cite{ai-opendata-tutorial} gives a tutorial
focusing on value of new AI insights in a domain, availability of relevant open data and context of people's interaction with the (new) system.
Consider water as an example. People make many daily decisions concerned with water involving
profession (e.g., fishing, irrigation, shipping), recreation (e.g, boating), wild life conservation (e.g., dolphins) or just regular living (e.g., drinking, bathing, washing). If accessible tools were available to public, they would be  particularly useful to handle public health challenges such as the Flint water crisis \cite{flint}. The very few tools available today target water experts such as WaterLive mobile app for Australia \footnote{http://www.water.nsw.gov.au/realtime-data
},  Bath app for UK\footnote{https://environment.data.gov.uk/bwq/profiles/}, and GangaWatch for India \cite{gw-demo} and assume technical understanding of the science behind water quality indicators. 
 
Psychologists have long  explored the sense-making process of analysts when looking at data through cognitive task analysis of their activities \cite{sensemaking-data}. 
They found that analysts try to explore schema and other metadata, look at data, build hypotheses and look for evidence. 
The current  interfaces to access data are intended for developers and data analysts. They consist of search interfaces on repository sites (like data.gov) or via search engines \footnote{https://toolbox.google.com/datasetsearch}; visualization\footnote{See examples at http://data.gov, http://data.gov.in.} and application programming interfaces (APIs). 
Further, for data analysis, analysts want  to understand the context of data available, the standards
and issues prevailing in a domain of their interest and a forum to discuss inter-disciplinary challenges. 
In fact, in \cite{chatbot-data-analysis}, the authors have created a chatbot to help with data analysis steps. 



But our focus is on how general public may benefit from insights generated from open data without long development cycles and in context of their use. 
The proposed chatbot generators will  enable a new, complimentary,
conversational, interface to open data that a user can interact with, possibly as part of a multi-modal user experience. Thus, a user will be able to talk to an automated chatbot to get insight she wants while optionally also seeing additional, relevant, visualizations and documents the agent may be able to retrieve. 



\section{Conversation Interfaces}
\label{sec:back}


\begin{figure*}
 \centering
   \includegraphics[width=0.65\textwidth]{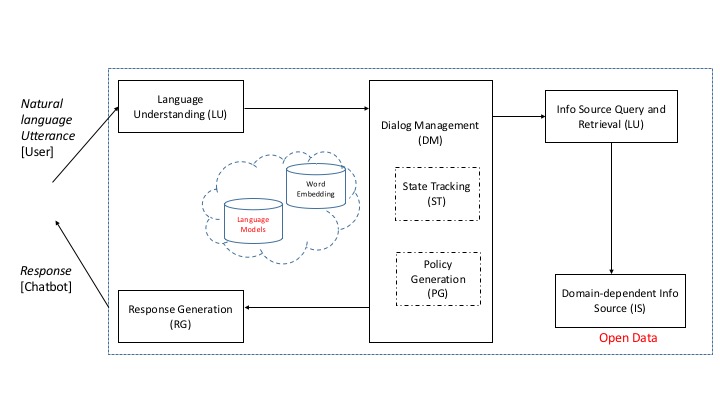}
  \caption{The architecture of a data-driven chatbot.}
  \label{fig:chatbot-arch}
\end{figure*}



There is  a long history of Conversational interfaces (CIs) going back to the 1960s when they first appeared to do casual conversation or answer questions \cite{dialog-intro}. 
A conversation, or {\em dialog}, is made up of a series of {\em turns}, where each turn is a series of {\em utterances} by one or more participants playing one or more {\em roles}. 
A common type of chatbot deals with a single user at a time and conducts informal conversation, answers the user's questions or provides recommendations in a given domain.
It needs to handle uncertainties related to human behavior and natural language, while conducting dialogs to achieve system goals. 

\subsection{Building Data-Consuming Chatbots}
\label{sec:builders}


The core problem in building chatbots is that of dialog management, i.e., creating dialog responses to user's utterances.
The system architecture of a typical data-consuming dialog manager  (DM) is shown in Figure~\ref{fig:chatbot-arch}. Given the user's utterance, it is analyzed to detect their intent and a 
policy for response is selected. This policy may call for querying a database, and the result is returned which is used by response generator to create a response using templates. The system can dynamically create one or more queries which involves selecting tables and attributes, filtering values and testing for conditions,  and assuming defaults for missing values. It may also decide not to answer a request if it is unsure of a query's result correctness.

Note that the DM may use one or more domain-specific data bases (sources) as well as one or more domain-independent sources like language models and word embeddings.
Common chatbots use static domain-dependent databases 
like product catalogs or user manuals. The application scenarios become more  compelling when the chatbot works in a dynamic environment, e.g., with sensor data, and interacts with groups of people, who come and go, rather than only an individual at a time. In such situations, the agent has to execute actions to monitor the environment, model different users engaged in conversation over time and track their intents, learn patterns and represent them, reason about best course of action given goals and system state, and execute conversation or other multi-modal actions. 


There are many approaches to tackle DM  in literature including finite-space, frame-based, inference-based and statistical learning-based \cite{chatbot-survey-statistical-ml,chatbot-book,minim-dialog,young2013pomdp}, of which,  finite-space and frame-based are most popular with mainstream developers. 
Task-oriented DMs  have traditionally been built using rules.
Further, a DM contains several independent modules which are optimized separately, relying on huge amount of human 
engineering .
The recent trend is to train DM from end-to-end, allowing error signal from the end output of DM to be back-propagated to raw input, so that the whole DM can be jointly optimized \cite{e2e-dialog-learning}.
A recent paper reviews the state-of-art and looks at  requirements and design options to make them customizable for end users as their own personal bot \cite{chatbot-arch-icse18}.

There are also unique considerations when exploring data with dialog:
\begin{itemize}
\item Dynamic source selection: the data in a domain may consist of multiple tables, attributes (columns) and rows. The user utterance could lead to discovering them and they then become part of the context of query. 
\item Query patterns: there are often common patterns and natural order of user queries. Users may adopt them to explore data and can be used as a shared context. 
\item Query cost: the order of query execution could be important for cost reasons. 
\item Query mapping: mappings from natural language to data model may have to be learned and adapted based on different source models.
\item Conversation length: As conversations become long, there could be increased risk of the user dropping off leading to diminishing returns.

\end{itemize}

\subsection{Usability Opportunities and Issues with Chatbots}

An obvious question to ask when considering chatbots is when they are most suitable. The effectiveness of conversation versus other modalities has long been studied \cite{direct-manipulation}. Some scenarios where conversation is quite suitable include when users are unfamiliar with the domain, expect non-human actors due to unique form of agent embodiment (e.g., robots) or prefer  them (e.g., due to sensitivity of the subject matter), and where content changes often and users seek guidance \cite{dialog-suitability}. 


However, such systems can also be fraught with ethical risks. An extreme and anecdotal example was the Tay \cite{tay} system in 2016 that was designed to engage with people on open topics over Twitter and learn from feedback, but ended up getting manipulated by users to exhibit unacceptable behavior via its extreme responses. 
The authors in \cite{dialog-ethics}  systematically  identify a number of potential ethical issues in dialogue systems built using learning methods:
showing implicit biases from data, being prone to adversarial examples, vulnerable to violate  privacy, need to maintain safety
of people, and concerns about explainability of response and reproducibility of results.
To handle these concerns, more research is needed. One 
idea here is to augment open data repositories that usually consist of standard information like data and size, 
usage license (context), how data was obtained (provenance),
semantics of missing values and units, and a responsible person.  Researchers have proposed to further describe protected variables and fairness considerations as datasheets \cite{dataset-datasheets}. The metadata can be used along with recent techniques and tools\footnote{AI Fairness Toolkit - https://github.com/IBM/AIF360} to address issues of fairness and transparency with chatbots \cite{dialog-ethics} and build trust with users.

\section{On Use-Cases With Open Data And Dialogs}
\label{sec:use-cases}

Over the years, a number of common analysis patterns have emerged across domains which have been shown to be useful to people around the world \cite{ai-opendata-tutorial}.  
One pattern  is {\em Return of Investment}. 
Here, the monetary investment made into a domain is compared
against suitable metrics of outcome and improvement is sought. As example, public funds invested into water work may be analyzed to see reduction in cases of heavy metal contamination, health care may be analyzed to see reduction in number of patients and deaths, or funds invested for tourist promotion be compared with  increase in economic activity in a city. Another pattern is {\em Comparison of Results}. Here, if an improvement is found in one context like domain, region or time, the user is interested to find whether the improvement holds for another similar context. Another pattern is {\em Demand-Supply Mismatch} where demand of a service/ resource like emergency visit in a city is compared with supply like health-care professionals. 
These could be good starting points 
for automated chatbot generators while technical experts 
focus on complex domain-dependent decision situations. 

Now consider a complex decision scenario. 
To make a decision on water consumption, 
one needs to consider the activity (purpose) for water use; relevant water quality parameters and their applicable regulatory standards for safety; available measurement technology, process, skills and costs; and actual data. 
These thus offer opportunities to integrate data from multiple sources and reason in the context of person's interest.
There are further complicating factors: there may be overlapping regulations due to geography and administrative scope; one may have to account for alternative ways to measure a particular water quality parameter that evolves over time; and water data can have issues like missing values or different levels of granularity. 
Therefore, use-cases like water  which do not follow 
common patterns
need to be tackled outside of the generator.  
 




\subsection{AI Methods in Multi-Modal Conversation - Water and Beyond}

\begin{figure}
 \centering
   \includegraphics[width=0.45\textwidth]{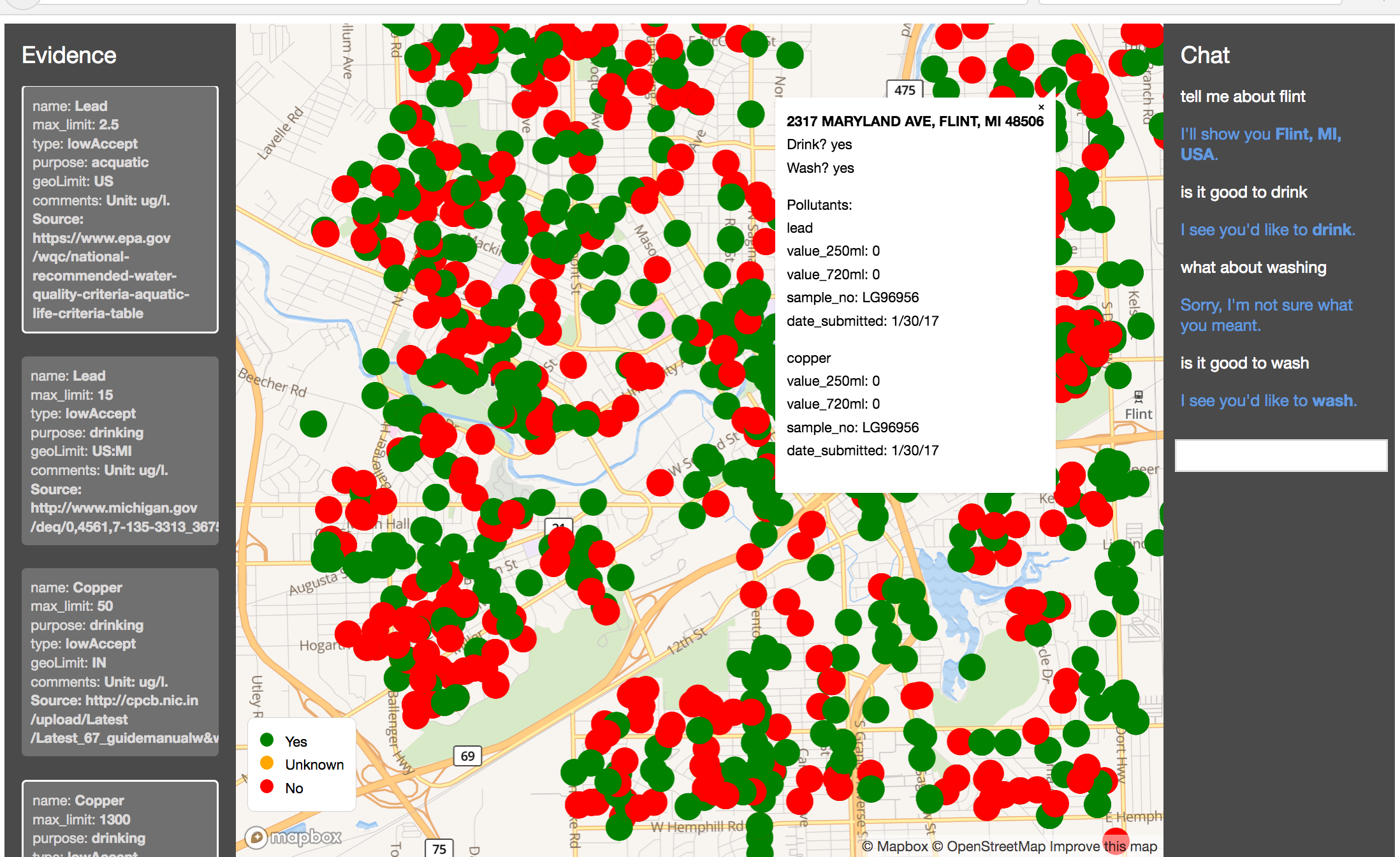}
  \caption{A screenshot of {\em Water Advisor}. See video of it in action at https://youtu.be/z4x44sxC3zA.}
  \label{fig:screen-shot-wa}
\end{figure}



We now illustrate how different AI methods came together to build a multi-modal chatbot like Water Advisor (WA) \cite{wa-aaai18}
and highlight how they would be relevant for 
a chatbot generator that works on common usage patterns.
There are also challenges for generalization which create 
opportunities for further research.
WA is intended to be a  data-driven assistant that can guide people globally without requiring any special water expertise.  One can trigger it via a conversation to get an overview of water condition at a location, explore it by filtering and zooming on a map, and seek details on demand (Figure~\ref{fig:screen-shot-wa}) by exploring relevant regulations, data or other locations.

\subsubsection{Learning} plays an important role in understanding user's utterances, selecting reliable data sources and improving overall performance over time. Specific to water, it is also used to  discovering issues in water quality together with regulation data. More generally, learning can be used for alternative DM approaches like end-to-end policy learning from data \cite{e2e-dialog-learning}.

\subsubsection{Representation} is needed to model location, time
 and user. For water, it encodes regulation and safe limits
 and mapping of usage purpose to quality parameters.

\subsubsection{Reasoning} is crucial to keep conversation focused based on system usability goals and user needs.  One can model cognitive costs to user based on alternative system response choices and seek to optimize short-term and long-term behavior. For water, reasoning is used to short-list regulations based on water activity and region of interest, generate advice and track explanations. 

\subsubsection{Execution} is autonomous as the agent can choose to act
by (a) asking clarifying questions about user intent  or locations, (b) asking user's preference
about advice, (c) seeking most reliable data source (water) for region and time interval of interest from available external data sources, and corresponding subset of compatible other sources (regulations) (d) invoking reasoning to generate an advice (for water usage using filtered water data and regulations), (e) visualizing its output and advice, and (f) using one or more suitable modalities available at any turn of user interaction, i.e., chat, maps and document views. 

\subsubsection{Human Usability Factors} have to be explicitly modeled and supported. In WA, the controller module for user-interface automatically keeps the different modalities synchronized and is aware of missing data or assumptions it is making, so that they can be used in
system response.
Further extensions can be to measure and track complexity of interaction \cite{dial-comp} and use sensed signals to pro-actively improve user experience and combine close-ended and open-ended questioning strategies for efficient interaction \cite{open-close-iui18}.

\subsubsection{Ethical Issues} can emerge whenever a piece of technology is used among people at large. We discussed handling them in a domain independent manner earlier. 
There may also use-cases needing  domain-specific considerations due to which scope of chatbot generator has to be selectively expanded. 




\section{Conclusion}
\label{sec:conc}

In this paper, we proposed the challenges of {\em intelligent conversation interface generator} that,  given a set of open data sources of interest, would generate a chatbot that can interact autonomously with a common person (non-developer) and provide insights about selected data. Such a technology would 
bring Artificial Intelligence (AI)-based solutions for important societal problems to the masses along common patterns of usage while technical AI experts focus on specialized cases. It will also serve as an  integrative testbed for human-centric AI and advance AI sub-areas. 

\bibliographystyle{named}
\bibliography{references}

\end{document}